\documentclass[lettersize,journal]{IEEEtran}
\usepackage{amsmath,amsfonts}
\usepackage{algorithmicx}
\usepackage{algorithm}
\usepackage{algpseudocode}
\usepackage{changes}
\usepackage{multirow}
\usepackage{array}
\usepackage[caption=false,font=normalsize,labelfont=sf,textfont=sf]{subfig}
\usepackage{textcomp}
\usepackage{stfloats}
\usepackage{url}
\usepackage{verbatim}
\usepackage{graphicx}
\usepackage{cite}
\usepackage{xcolor}
\usepackage{color}
\usepackage{array, makecell}
\usepackage{amssymb}
\usepackage{pifont}
\usepackage{threeparttable}
\usepackage{booktabs}
\usepackage{array}
\usepackage{makecell}

\begin{document}

\title{Adaptive Soft Error Protection for Neural Network Processing }

\author {Xinghua Xue, ~\IEEEmembership{Member,~IEEE},
         Cheng Liu, ~\IEEEmembership{Senior Member,~IEEE},
         Feng Min, \\
          Hui Li, 
          Kai Zhang, ~\IEEEmembership{Member,~IEEE},
         Yinhe Han, ~\IEEEmembership{Senior Member,~IEEE}


\thanks{Cheng Liu is the corresponding author. Xinghua Xue is with Hangzhou Institute for Advanced Study, University of Chinese Academy of Sciences. Cheng Liu is with the State Key Lab of Processors, Institute of Computing Technology, Chinese Academy of Sciences. Feng Min is with Institute of Computing Technology, Chinese Academy of Sciences. Hui Li and Kai Zhang are with Jinan Inspur Data Technology Co., Ltd. Yinhe Han is with Institute of Computing Technology, Chinese Academy of Sciences. (e-mail:xuexinghua@ucas.ac.cn)}

\thanks{This paper is supported in part by National Natural Science Foundation of China (NSFC) under grant No.(U24A20291) and National Natural Science Foundation of China (NSFC) under grant No.(62174162).}
}

\markboth{IEEE Transactions on Computer-aided Design of Integrated Circuits and Systems, ~Vol.~xx, No.~xx, xxx~2026}%
{Xinghua Xue \MakeLowercase{\textit{et al.}}: Bare Demo of IEEEtran.cls for IEEE Journals}

\maketitle

\begin{abstract}

Previous research on selective protection for neural network components typically exploits only static vulnerability differences. Although these methods improve upon classical modular redundancy, they still incur substantial overhead for neural network workloads that are both memory-intensive and compute-intensive. In this work, we observe that neural network vulnerability is also input-dependent and varies dynamically at runtime. With this observation, we propose an adaptive, vulnerability-aware fault tolerance framework. At its core, a lightweight graph neural network (GNN) model dynamically predicts soft error vulnerabilities across inputs and neural network components, enabling real-time adaptation of fault tolerance policies. This design offers a complementary and more efficient protection scheme compared to traditional approaches. Experimental results demonstrate that the GNN predictor achieves over 95\% accuracy in identifying critical inputs and components. Moreover, our adaptive scheme reduces  computational  overhead by an average of 42.12\% while preserving model accuracy, significantly outperforming static selective protection methods.

\end{abstract}


\section{Introduction}
Recent advancements in deep learning have significantly impacted sectors such as aerospace, autonomous driving, and medical robotics\cite{zaoui2024impact, rane2024emerging, cui2024survey}. These fields are usually sensitive to faults and necessitate reliable computational processing \cite{perez2024artificial, prokop2024software}. Nonetheless, the ongoing miniaturization of chips and increasing integration due to semiconductor manufacturing technology improvements have led to a surge in the occurrence of soft errors, despite the associated gains in energy efficiency. The capricious nature of these soft errors, coupled with their propensity to remain concealed and propagate, poses a significant threat to the reliability of applications. For safety-critical applications with high reliability requirements, minor disturbances can even cause catastrophic consequences\cite{neggaz2018reliability}. Consequently, the incorporation of fault-tolerant techniques in deep learning has become essential to sustain system integrity and continuous and reliable operations.

Previous studies\cite{sun2025ft2,rajappa2023smart, ozen2025linear,li2017understanding,caro2025semantic,rech2024artificial,  sharif2023efficient} have examined the reliability of deep learning models and proposed a variety of software-based approaches to address soft errors, including triple module redundancy (TMR), dual module redundancy (DMR), and algorithm-based fault tolerance (ABFT). Given deep learning's substantial need for computational and memory capacity, classical fault-tolerant techniques have been noted to significantly increase computational overhead.

To this end, researchers have investigated the different vulnerabilities of various computational engine components and neural network structures through  simulation-based fault injection approaches, with the aim of implementing protection for the most vulnerable parts to minimize protection overhead\cite{ bertoa2022fault,hsieh2023cost,koylu2022smart,xue2022winograd,weng2024fkeras,cai2024evaluation}.  For example, some  have identified the distinct vulnerabilities across neural network layers, preferring to safeguard the most vulnerable layers to lower fault tolerance costs. The authors in \cite{xu2021reliability} noted that deep learning accelerators' control-intensive modules, such as the address generator and instruction decoder, are particularly susceptible to hardware faults, and have effectively applied TMR to these modules, achieving significant improvements in reliability with minimal hardware overhead. Furthermore, studies in neural network quantization\cite{taheri2024exploration} also exploit variations in sensitivity to quantization errors, adopting bespoke quantization strategies for different neural network segments. These strategies have proven the substantial benefits of adopting selective fault-tolerant designs.

However, as neural network models become more complex and have higher error rates, the overhead of exploring vulnerability differences and implementing selective fault-tolerant protection through simulation-based fault injection techniques increases significantly. In addition, existing studies perform a single-dimensional and static analysis of vulnerability differences among neural network components. These schemes ignore the vulnerability coupling relationship between different inputs and neural network components. This ignores the dynamic impact of different inputs on the vulnerability of neural networks.

Unlike static vulnerability analysis approaches, we observe that in addition to the model layer, the vulnerability of neural networks is highly influenced by input data and varies dynamically at runtime. Specifically, simpler deep learning tasks tend to be more resilient to soft errors, making it inefficient to apply intensive fault protection uniformly across all inputs and layers. To this end, we propose an adaptive fault-tolerant design framework that dynamically adjusts protection intensity based on the varying vulnerabilities of both neural network layers and input data, providing important opportunities for optimizing fault-tolerant methods.

 To accurately capture the input and component-specific soft error vulnerability and enable the runtime vulnerability prediction, we introduce a lightweight graph neural network (GNN) model. Since the forward computation process of a neural network can essentially be viewed as a directed computational graph, the impact of soft errors propagates and accumulates along the dependency edges within the graph. The GNN aggregates neighborhood information, explicitly capturing the complex interactions between different components and the global impact of fault propagation. This overcomes the limitations of traditional machine learning methods, which extract features only from isolated components and struggle to represent structural dependencies. Furthermore, compared to CNN designed for regular grids, GNN can handle computational graph structures more efficiently and naturally. Therefore, GNN is suitable for modeling and predicting dynamic vulnerability effects arising from the coupling of input data and internal computational structures.  Experimental results demonstrate that our approach can seamlessly integrate with existing fault-tolerant techniques, significantly reducing computational overhead while maintaining fault tolerance capabilities.

The main contributions are summarized as follows.

\begin{itemize}

\item We analyze input vulnerability and identify significant disparities in responses to soft errors, paving the way for adaptive fault-tolerant strategies.

\item We propose a lightweight GNN model that effectively captures input- and component-specific soft error vulnerabilities while enabling real-time vulnerability prediction.

\item Building on the vulnerability predictor, we develop an adaptive fault-tolerant strategy that dynamically adjusts protection intensity based on the predicted vulnerability of each neural network input and component.

\item Experimental results on various datasets and neural networks demonstrate that the proposed adaptive strategy, integrated with a classical fault-tolerant approach, reduces computational overhead by an average of 42.12\% without compromising reliability, compared to static vulnerability-based approaches.

\end{itemize}

\section{RELATED WORK} 

The underlying computing engines for deep learning processing are typically fabricated using silicon-based nanotechnology, such as neural network accelerators, GPUs, and CPUs. These components inevitably suffer from high-energy neutrons, manufacturing defects, aging, temperature, etc. Such factors induce soft errors, which can cause unpredictable data corruption or system failures\cite{baumann2005soft,ziegler2004ser}. Furthermore, the continued scaling down of semiconductor manufacturing processes leads to steadily increasing integration density in integrated circuit chips\cite{borkar2006designing}. This trend amplifies the failure risk for these chips. Hence, fault-tolerant design becomes essential.

Traditional fault tolerance schemes based on full redundancy incur high overheads. To mitigate this, researchers have proposed selective fault tolerance protection using simulation-based fault injection approaches. For example, Ordoñez et al. \cite{ordonez2024enhancing} protected the most vulnerable layers in neural network models against targeted fault injection attacks. Ahmadilivani et al.\cite{ahmadilivani2024cost} implemented selective replication of filters/neurons through parametric vulnerability analysis. This enables efficient correction of output channels via robust correction layers, while also applying cost-effective pruning techniques based on parameter vulnerability. Nazari et al.\cite{nazari2025reliability} developed a hybrid method combining Genetic Algorithms (GA) with fault injection for rapid and accurate layer vulnerability analysis in quantized DNNs, applying layer-specific quantization levels. These simulation-based selective fault tolerance methods effectively enhance neural network reliability. However, when DNN models are more complex and have higher fault rates, the simulation time and computation required increase dramatically.

To address this challenge, researchers have proposed machine learning techniques for reliability analysis, aiming to provide a basis for subsequent selective fault-tolerant design. For example, Lange et al.\cite{lange2020machine} utilized a machine learning clustering algorithm to efficiently group triggers with similar functional failure sensitivities. Based on the clustering results, failure simulation activities can be performed in groups, significantly reducing the computational and time overhead. However, existing machine learning-based reliability analysis methods typically focus on extracting and learning features from individual components. These methods fail to adequately model fault propagation paths or consider their global impact.

GNNs have a natural topology modeling capability to explicitly capture the complex dependencies and fault propagation mechanisms of different system components. This provides a new research direction to address the above limitations. For example, Liu et al.\cite{liu2023fern} proposed a GAT-based architecture to model network topology, traffic demand, routing, and target failure  scenarios to predict the impact of link failures. Karim et al.\cite{karim2025toward} employed GNNs to model nanosatellite subsystem interactions, enabling real-time detection of Single-Event Effect (SEE)-induced faults and anomalies. Lu et al.\cite{lu2025accelerate,lu2024machine,lu2023toward} used GNNs to learn circuit-level fault propagation patterns, achieving accurate prediction of flip-flop fault sensitivity. Koufopoulou et al.\cite{koufopoulou2023prediction} applied GNNs to High-Level Synthesis (HLS) circuit security assessment, enabling efficient quantification of hardware security attributes. Ros et al.\cite{ros2024handling} integrated GNNs with Deep Reinforcement Learning (DRL) to enhance fault tolerance in large-scale IoT service function chains. Although GNNs have proven effective in reliability analysis applications, they remain unexplored for capturing dynamic vulnerability effects arising between neural network inputs and internal components.

\section{Motivation}

Many prior works have already investigated the soft error vulnerability variations across different layers of neural network models for selective protection.  However, inputs that can either be simpler or complex for typical neural network tasks such as classification and object detection are generally overlooked. In this section, we mainly explore the vulnerability variations between different tasks when faults are randomly injected into different inputs and model layers, which reveals the potential benefits of adaptive protection against soft errors.

\begin{figure}[!h]
\centering

\includegraphics[width=0.45\textwidth]{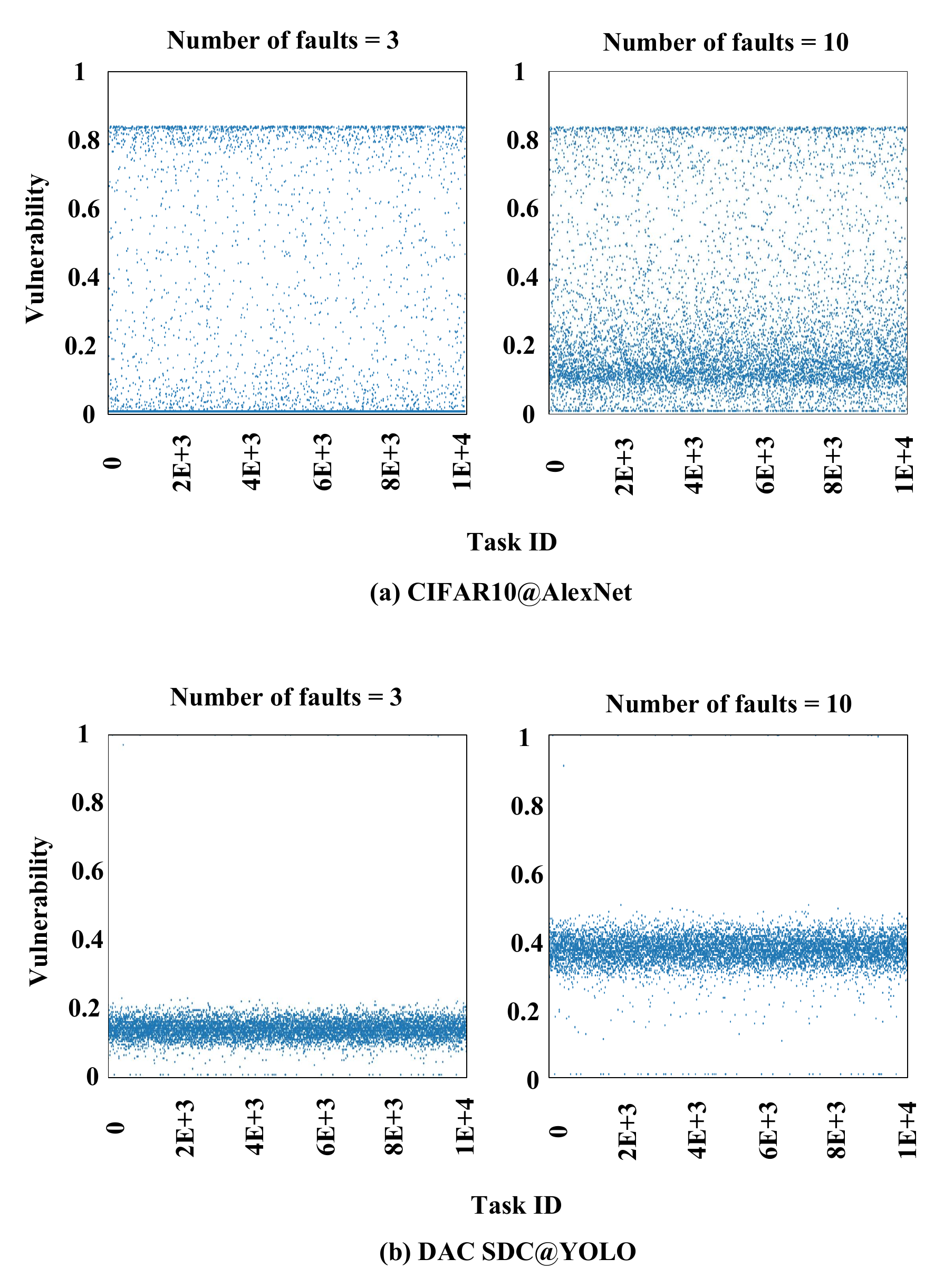}

\caption{ Vulnerability variations between different tasks when faults are randomly injected into different inputs and model layers.}
\label{fig:img1}
\end{figure}

We select two representative cases for detailed analysis: the image classification task of the AlexNet model on the CIFAR-10 dataset, and the YOLO object detection model on the autonomous vehicle dataset from the DAC System Design Contest (SDC)\cite{xu2019dac}. We quantify the soft error vulnerability of each model. Specifically, for each task, the difference between the accuracy of the error-free baseline model and the accuracy after injecting soft errors is calculated as the final vulnerability value. The experiment simulates the soft error effect by randomly selecting bits from the input image and model layers for state flipping. Fig. \ref{fig:img1} illustrates the experimental results, where the horizontal axis represents the task sample ID and the vertical axis quantitatively represents the soft error vulnerability, with larger data values indicating greater vulnerability.

Experimental results demonstrate that even when faults are randomly distributed throughout the entire network, different input tasks still exhibit significant variations in vulnerability. The vulnerability levels of some input task samples are very high, indicating that they are extremely sensitive to soft error perturbations. Therefore, it is necessary to deploy enhanced fault-tolerant protection mechanisms to maintain system reliability. In contrast, the vulnerability characteristics of some input tasks are negligible, and their  vulnerability values are close to 0. At the same time, the vulnerability distributions differ significantly across different model architectures and datasets. For example, the AlexNet model trained on the CIFAR-10 dataset exhibits a discrete vulnerability distribution, while the YOLO model running on the DAC SDC dataset has a relatively concentrated vulnerability distribution. In addition, the number of injected soft errors exerts a decisive influence on the vulnerability distribution. In general, the larger the number of injected errors, the higher the overall vulnerability level of the model and the more discrete the vulnerability distribution. Therefore, in scenarios with high error rates, more comprehensive fault-tolerant designs need to be deployed.

In summary, this experiment reveals considerable variability in vulnerability across various inputs and models.  To circumvent the issues of excessive computational burden and diminished accuracy contributions from over-protection, adaptive fault-tolerant strategies should be tailored to the complexity of the inputs and models.

\section{Adaptive vulnerability-aware Fault-Tolerant Design Framework}

\begin{figure*}[!t]
\centering
\vspace{-1.5em}
\includegraphics[width=1\textwidth]{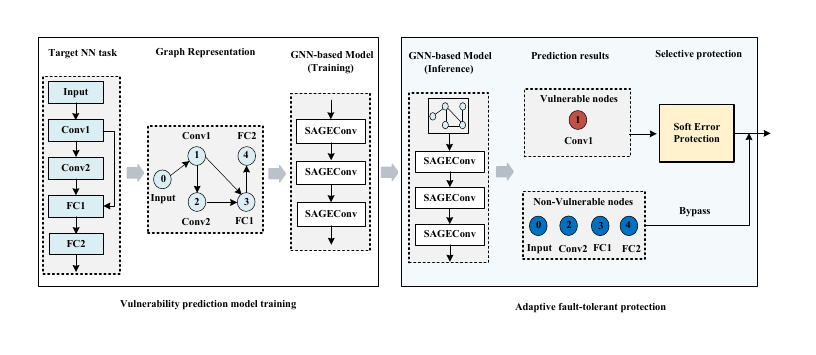}
\vspace{-2.5em}
\caption{The proposed adaptive fault-tolerant design framework. It leverages a GNN model to predict neural network vulnerability to soft errors. The prediction is further utilized to determine whether a neural network layer is vulnerable to errors and requires intensive protection at runtime.}
\label{fig:img2}
\end{figure*}

\subsection{Framework Overview}
By leveraging the substantial variations in neural network vulnerability, selective  protection can focus on safeguarding the more error-prone regions while avoiding excessive protection of less vulnerable parts, thereby reducing overall overhead. However, most existing selective protection approaches rely on static vulnerability assessments of the network architecture, derived through offline analysis, which remain fixed during runtime. As a result, these approaches fail to adapt to the dynamic vulnerability variations caused by different inputs. Notably, the vulnerability across the neural network architecture is closely influenced by input-dependent factors. To address this challenge, a dynamic input- and layer-aware vulnerability predictor is essential for effective selective protection. Such a predictor must not only account for accurate vulnerability prediction but also be lightweight enough to operate efficiently at runtime.

The adaptive input and layer vulnerability-aware fault-tolerant design framework consists of two main parts, illustrated in Fig. \ref{fig:img2}. Initially, a GNN-based predictor is utilized to efficiently predict task vulnerability. Following this, a fault tolerance strategy is dynamically implemented based on the predicted task vulnerability, aiming to optimize the fault tolerance. Given that some input tasks and layers of neural networks exhibit fault insensitivity, fault tolerance may be unnecessary. To further improve fault tolerance performance, we consider both input vulnerability and layer vulnerability.

Specifically, we propose an efficient and lightweight GNN-based vulnerability predictor to accurately assess neural network vulnerabilities.  As illustrated in the left part of Fig. \ref{fig:img2}, the predictor models the entire neural network including both inputs and layers as a computational graph and embeds each node as a vector for accurate vulnerability prediction. To simplify the prediction process for efficient runtime deployment, we categorize nodes into two groups. Then, the model is trained offline using the constructed vulnerability dataset. When the GNN-based vulnerability prediction model is ready, it is applied to each input and layer, and then adaptive fault tolerance protection is performed, as shown in the right part of Fig. \ref{fig:img2}. Each node is classified, and protection is applied exclusively to vulnerable nodes, enabling resource-efficient fault tolerance.

\begin{figure}[!t]
\centering
\vspace{-2em}
\includegraphics[width=0.48\textwidth]{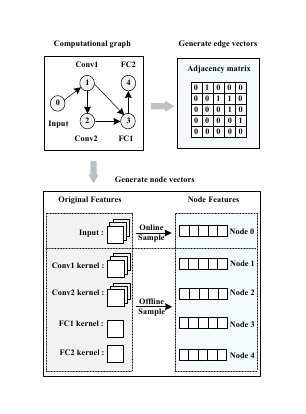}
\vspace{-4em}
\caption{An example of graph representation.}
\label{fig:img3}
\end{figure}

\subsection{GNN-based Vulnerability Predictor}

Our goal is to perform node classification tasks on the constructed graph, utilizing the powerful capabilities of GNN to identify vulnerable inputs and layers within the neural network model. The proposed predictor is framed as a standard node classification problem and incorporates three SAGEConv layers \cite{hamilton2017inductive}. Each node is classified into one of two output labels: vulnerable or non-vulnerable, making the overall model lightweight and efficient. The GNN-based vulnerability prediction model is critical to the proposed adaptive protection framework. Its prediction accuracy mainly depends on the graph construction and the node vectorization. 

\subsubsection{Graph Construction}  In order to characterize  the influence of both inputs and network architecture on neural network vulnerability, both inputs and the neural network components are modeled as nodes in a graph $G(V,E)$, while the direct computational dependencies between these nodes are modeled as edges. Note that $V$ and $E$ denote the node set and edge set of the graph respectively. In this work, we take each layer as a basic component or node for the simplicity of representation, but it can either be combined or split to suit the most appropriate protection granularities. Since activations are essentially intermediate results of the neural network processing and they are only known at runtime, we utilize weights in each layer to construct the node vectors.

\begin{figure*}[!t]
\centering

\includegraphics[width=1\textwidth]{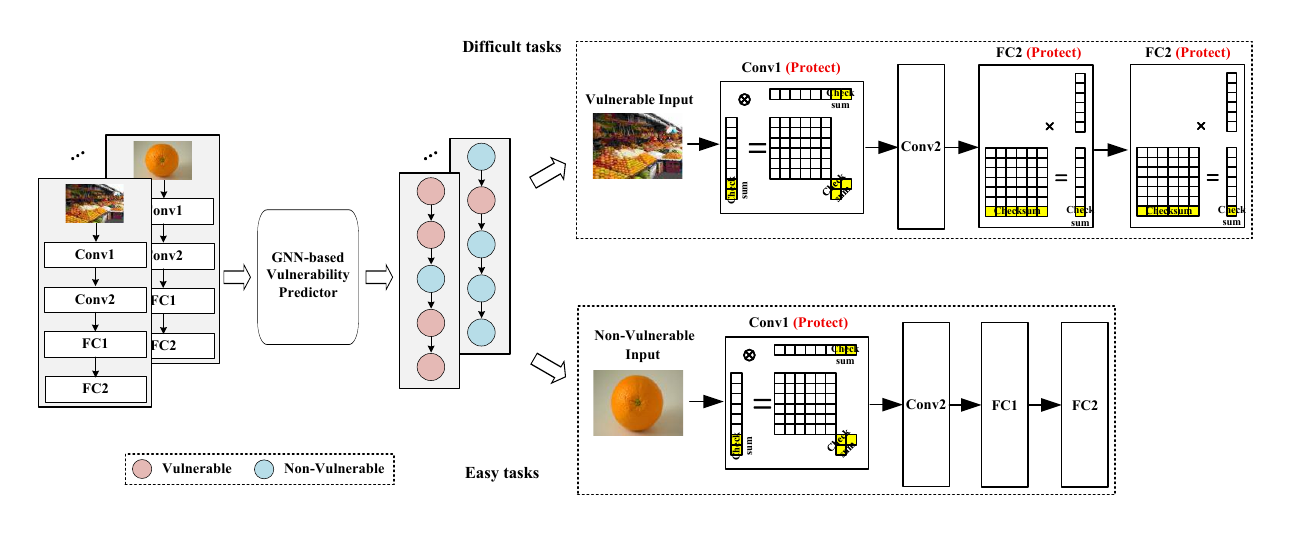}
\vspace{-2em}
\caption{An ABFT example enhanced with the proposed adaptive fault-tolerance protection.}
\label{fig:protect}
\end{figure*}

Fig. \ref{fig:img3} exemplifies the graph representation of a typical neural network model. The neural network model includes an input image, two convolutional (Conv) layers, and two fully connected (FC) layers, resulting in a five-node graph. For GNN-based vulnerability prediction, each node of the graph is encoded as a feature vector. However, as neural network inputs and layers are often high-dimensional, directly converting raw data into large feature vectors can be computationally inefficient. While various approaches, such as principal component analysis (PCA) and neural network-based embedding models, have been proposed for the feature vector construction, these approaches are typically time-consuming, making them unsuitable for runtime predictions. In this work, we use a downsampling method based on max pooling to extract feature vectors from high-dimensional data as node vectors. The dependencies between nodes are encoded as an adjacency matrix, where 1 indicates the existence of a directed edge between nodes.

\subsubsection{Vulnerability dataset construction}
To train the vulnerability predictor, we construct a dataset comprising feature vectors of tasks and their corresponding vulnerability labels.  By re-labeling existing open-source datasets and different layers of deep learning models (such as convolutional layers and fully connected layers), vulnerability labels reflecting the sensitivity of tasks to errors are assigned. To improve the efficiency of subsequent vulnerability prediction models, the complex vulnerability level is simplified into a binary classification system, i.e., the vulnerable and the non-vulnerable.

The dataset label generation process comprises three key stages. First, based on predefined fault injection parameters, one or more bits in the input image data or the output of arithmetic operations in the model layer are randomly flipped to simulate a fault interference scenario. Then, forward inference is performed only when a fault is injected into the test image or model layer, and the model output accuracy is recorded. The difference between this accuracy and the accuracy under fault-free baseline conditions is calculated as the final vulnerability value. Finally, based on the vulnerability value distribution of the tasks, the average vulnerability of all tasks is calculated as the classification threshold. Tasks with vulnerability values below the average are labeled as non-vulnerable, while tasks with vulnerability values above the average are labeled as vulnerable. Notably, the entire label generation process is a one-time, offline operation.

\subsection{Adaptive Fault-tolerant Protection}

Based on the vulnerability prediction results, the framework applies dynamic protection strategies to balance protection strength against performance overhead. It integrates seamlessly with classical fault tolerance techniques such as triple modular redundancy (TMR) and algorithm-based fault tolerance (ABFT), serving as a complementary enhancement rather than a replacement. To achieve just enough fault tolerance, the protection intensity is aligned with the predicted vulnerability intensity of the neural network components. Most fault-tolerant approaches can be scaled to trade off protection coverage with overhead. For example, TMR can be applied at different levels of granularity, while ABFT for convolution and general matrix multiplication (GEMM) \cite{zhao2020ft, xue2023soft} can be tuned by adjusting checksum granularity or thresholds.

However, GNN predictor accuracy tends to degrade when vulnerability classification is made more fine-grained. To strike a balance, we categorize vulnerability into two levels, i.e., vulnerable and non-vulnerable, and demonstrate the adaptive protection scheme using ABFT as a representative case. As illustrated in Fig. \ref{fig:protect}, unlike uniform protection schemes that apply ABFT uniformly across all inputs and layers, the proposed method employs dynamic protection guided by the GNN predictor. Specifically, for inputs classified as non-vulnerable, a relaxed strategy is adopted to minimize overhead without sacrificing overall performance. In contrast, for inputs identified as highly vulnerable, stricter measures are selectively applied to ensure reliability. Furthermore, while considering the input task vulnerability, we jointly consider the vulnerability results across different network layers. This dual evaluation mechanism enables us to identify the most critical inputs and layers and apply targeted protection, thereby enhancing efficiency while maintaining robustness.

\begin{figure*}[!t]
\centering

\includegraphics[width=1\textwidth]{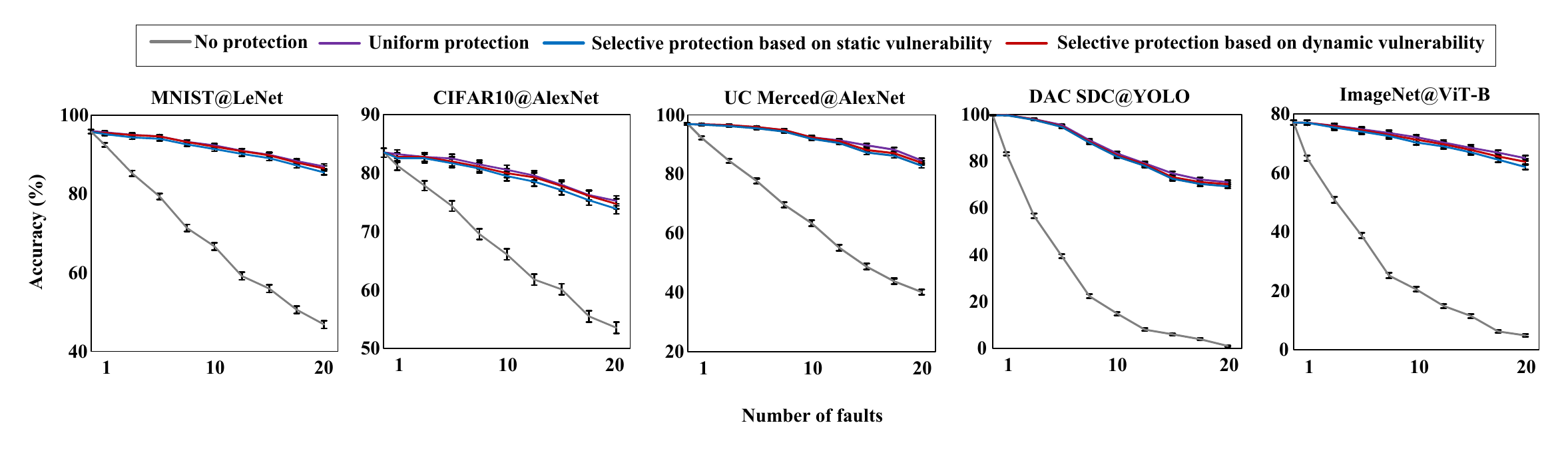}
\vspace{-0.5em}
\caption{Model accuracy comparison between different fault-tolerant design strategies under various fault injection setups. }
\label{fig:img9}
\end{figure*}

\section{Experiment results}

\subsection{Experiment Setups} 

\subsubsection{Datasets And Models} The following neural network models and datasets are selected as benchmarks for evaluation. LeNet-5 is trained on the MNIST dataset\cite{lecun1998gradient}, which contains 60,000 training images and 10,000 test images. AlexNet is trained on the CIFAR-10 dataset\cite{krizhevsky2009learning} containing 50,000 training images and 10,000 test images, as well as the UC Merced dataset\cite{yang2010bag} comprising 2,100 images, 30\% of which are test images. YOLOv3\cite{redmon2018yolov3} is trained on the  DAC SDC dataset\cite{xu2019dac}, which comprises 93,520 images, 10,000 of which are test images. ViT-B\cite{dosovitskiy2020image} is trained on the ImageNet dataset\cite{deng2009imagenet}, which uses 100,000 training images, 50,000 validation images, and 10,000 test images. The GNN-based vulnerability predictor consists of three SAGEConv layers. All neural network models are quantized to INT8. All models are trained using the PyTorch framework for 200 epochs with a batch size of 256.

\subsubsection{Evaluation Metrics} To comprehensively and accurately evaluate the proposed fault-tolerant method, we employ multi-dimensional evaluation metrics, including model accuracy, computational overhead, and runtime.  Model accuracy is measured using Top-1 classification accuracy. Computational overhead is defined as the additional arithmetic operations introduced by the fault-tolerant algorithm, quantified in terms of primitive operations such as additions and multiplications. The normalized computational overhead is calculated relative to the baseline fault-tolerant approach, which applies a uniform protection strategy across all input tasks and model layers without accounting for vulnerability differences. Runtime measures the execution time or latency of the proposed methods during inference.

\subsubsection{Fault Injection} In this work, we employ the previously proposed soft error fault injection simulation method \cite{huang2024mrfi} to inject faults into different input tasks and different layers in the neural network and perform vulnerability assessment. The fault injection mechanism is implemented within the PyTorch framework, utilizing its built-in hook mechanism to perform bit-level perturbation operations.  Specifically, random bit flips (0→1 or 1→0) are introduced in the input data stream of the neural network or the output results of core arithmetic operations of model layers to simulate soft errors. It is worth noting that MRFI is a general fault injection framework that abstracts away low-level hardware architectural details. Similar to instruction-level fault injection mechanisms \cite{tsai2021nvbitfi}, it operates at the granularity of network operators, such as convolution layers. While this design sacrifices some accuracy in fault injection, it achieves higher speed and broader applicability. As a result, MRFI is well-suited for evaluating the proposed adaptive fault-tolerant approach, which can be applied across diverse hardware architectures. The number of faults serves as a key metric, representing the total number of injected bit-flip errors.

To ensure the statistical  reliability of the evaluation results, we employ a statistical method based on the margin of error as the theoretical basis for our experimental design, which determines the minimum number of experiments required for a specific confidence level\cite{leveugle2009statistical}. According to relevant survey studies \cite{ahmadilivani2024systematic}, a 95\% confidence level and a ±2\% margin of error are typically adopted. Based on this, Table \ref{tab:min_trials1} lists the theoretical minimum number of experiments required for each model. To ensure the reliability of the results, all experiments were conducted 1,000 times, exceeding these theoretical minimum requirements. The reported values include the mean and its 95\% confidence interval, with the confidence interval visually represented by error bars.

\begin{table}[htbp]
\centering
\scriptsize

\setlength{\tabcolsep}{5pt}
\caption{Theoretical minimum number of experiments.}
\label{tab:min_trials1}
\begin{tabular}{lc}
\toprule
\textbf{Dataset@Model} & \makecell{\textbf{Minimum number} \\ \textbf{of experiment}} \\
\midrule
MNIST@LeNet        & 457 \\
CIFAR-10@AlexNet   & 948 \\
UC Merced@AlexNet  & 369 \\
DAC SDC@YOLO       &  343 \\
ImageNet@ViT-B     & 963 \\
\bottomrule
\end{tabular}
\end{table}

\subsubsection{Hardware Platforms} The evaluation experiments were performed on a server equipped with two 24-core@2.5GHz Intel Xeon processors, 512GB memory, and four PH402 SKU 200 GPU cards.

\subsection{Overall Fault-tolerant Design Evaluation}

To demonstrate the performance advantages of the proposed adaptive fault-tolerant design, we evaluate it from three perspectives: model accuracy, computational overhead, and runtime. It is compared with two fault-tolerant methods. First, the uniform protection method draws on the ABFT fault-tolerant algorithms for convolution and GEMM proposed in \cite {zhao2020ft} and \cite {xue2023soft} to implement an error protection mechanism across all input tasks and model layers.  Second, the selective protection method based on static vulnerability draws on the vulnerability analysis of each layer based on fault injection in \cite {xue2023soft} to provide targeted protection for key layers.

\subsubsection{Accuracy Evaluation}  Fig. \ref{fig:img9} compares the accuracy of our adaptive fault-tolerant approach against the baseline fault-tolerant approach and the static fault-tolerant approach under varying fault conditions. The experimental results reveal that, under conditions with a low number of faults, our adaptive fault-tolerant protection attains accuracy comparable to the no-fault scenario, thereby exhibiting its outstanding fault-tolerant capability. As the number of faults increases, our approach consistently achieves satisfactory results. Although the accuracy may decrease in some scenarios, the gain in accuracy is still very large. Our proposed approach attains near-identical accuracy to the baseline  protection, thereby validating its compatibility with classic fault-tolerant techniques without compromising the overall  effectiveness.

\subsubsection{Computational Overhead Evaluation}

\begin{table*}[htbp]
\centering
\scriptsize

\setlength{\tabcolsep}{15pt}
\renewcommand{\arraystretch}{1.2}
\caption{Comparison of Additional Computational Overhead (\# of arithmetic operations) Introduced by Different Fault-Tolerant Design Strategies Under Various Fault Injection setups.}
\label{tab:comp_comparison}
\begin{tabular}{@{}ll *{5}{c}@{}}
\toprule
\multirow{2}{*}{\textbf{Dataset@Model}} & \multirow{2}{*}{\textbf{Protection Method}} & \multicolumn{5}{c}{\textbf{Number of faults}} \\
\cmidrule(lr){3-7}
& & \textbf{1} & \textbf{5} & \textbf{10} & \textbf{15} & \textbf{20} \\
\midrule

\multirow{3}{*}{MNIST@LeNet} 
& Uniform Protection & 4.772e4 & 5.401e4 & 6.187e4 & 6.973e4 & 7.759e4 \\
& Static Protection & 2.894e4 & 4.291e4 & 4.953e4 & 6.035e4 & 7.018e4 \\
& Proposed Protection & 1.777e4 & 2.228e4 & 2.709e4 & 3.076e4 & 3.478e4 \\

\addlinespace
\addlinespace

\multirow{3}{*}{CIFAR10@AlexNet}
& Uniform Protection & 3.772e6 & 3.931e6 & 4.131e6 & 4.330e6 & 4.529e6 \\
& Static Protection & 1.114e6 & 3.27e6 & 3.904e6 & 4.145e6 & 4.36e6 \\
& Proposed Protection & 1.071e6 & 1.615e6 & 1.914e6 & 2.14e6 & 2.357e6 \\

\addlinespace
\addlinespace

\multirow{3}{*}{UC Merced@AlexNet}
& Uniform Protection & 1.098e7 & 1.113e7 & 1.132e7 & 1.151e7 & 1.17e7 \\
& Static Protection & 8.086e6 & 8.245e6 & 8.623e6 & 9.281e6 & 9.936e6 \\
& Proposed Protection & 1.475e6 & 3.234e6 & 4.723e6 & 5.81e6 & 6.451e6 \\

\addlinespace
\addlinespace

\multirow{3}{*}{DAC SDC@YOLO}
& Uniform Protection & 9.479e6 & 9.555e6 & 9.65e6 & 9.745e6 & 9.84e6 \\
& Static Protection & 6.532e6 & 7.425e6 & 7.871e6 & 8.548e6 & 9.236e6 \\
& Proposed Protection & 4.036e6 & 4.556e6 & 4.94e6 & 5.062e6 & 5.256e6 \\

\addlinespace
\addlinespace

\multirow{3}{*}{ImageNet@ViT-B}
& Uniform Protection & 1.049e8 & 1.052e8 & 1.056e8 & 1.06e8 & 1.064e8 \\
& Static Protection & 6.992e7 & 7.89e7 & 8.799e7 & 9.034e7 & 9.75e7 \\
& Proposed Protection & 3.755e7 & 4.843e7 & 5.311e7 & 5.737e7 & 6.397e7 \\

\bottomrule
\end{tabular}
\end{table*}

Table \ref{tab:comp_comparison} compares the fault-tolerant computational overhead of the proposed approach. Experimental results show that compared with the baseline uniform approach, the proposed approach significantly optimizes the baseline fault-tolerant protection, reducing the computational overhead by an average of 59.19\%. Compared with the static fault-tolerant approach, our approach reduces the computational overhead by 42.12\% on average without affecting the fault-tolerant capability. In addition, it is found that the reduction of computational overhead varies in different dataset tasks and fault conditions.

\begin{figure}[t]
\centering
\vspace{-1em}
\includegraphics[width=0.45\textwidth]{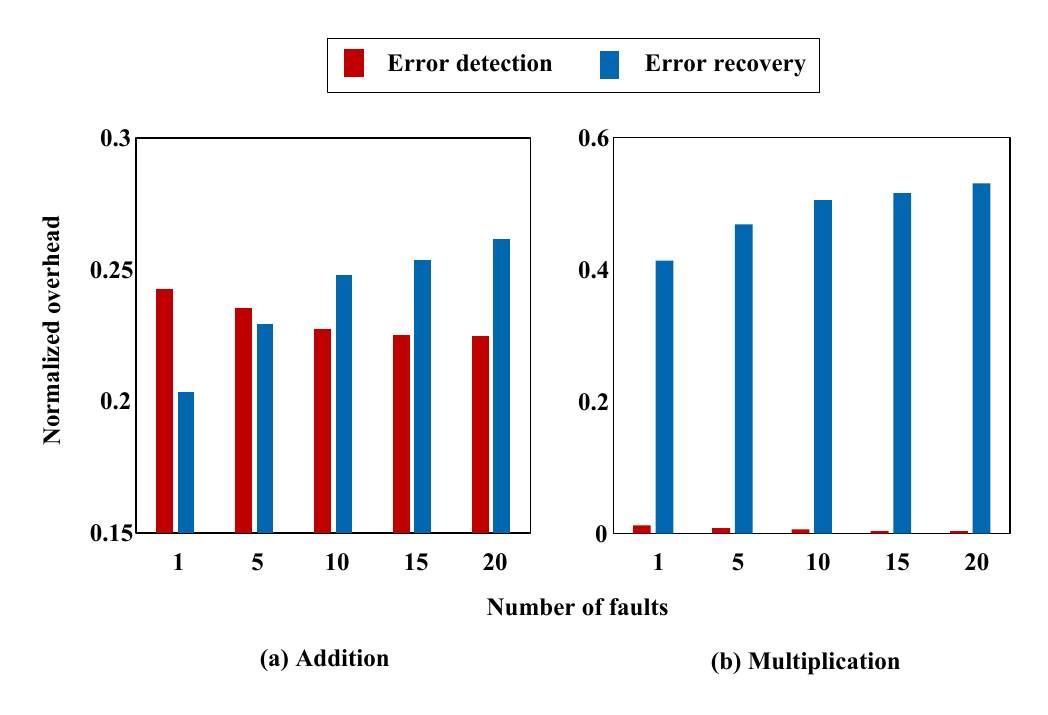}
\vspace{-1em}
\caption{Computational overhead of error detection and error recovery is evaluated using the YOLO model on the DAC SDC dataset, normalized to the uniform protection.}
\label{fig:comp_detec_recovery}
\end{figure}

For a deeper insight into the computational overhead of our proposed adaptive fault-tolerant strategy, Fig. \ref{fig:comp_detec_recovery} further analyzes the fault-tolerant design overhead of  error detection and  error recovery of ABFT in the addition and multiplication operations on the YOLO model of the DAC SDC dataset, respectively. The computational overhead is normalized to that of a uniform protection. It can be found that in the case of low faults, the overhead of the addition operation of error detection accounts for a larger proportion, and as the number of faults increases, the proportion of error recovery gradually increases. This analysis aims to provide a valuable reference for further performance optimization.

\subsubsection{Runtime Evaluation}

\begin{table*}[htbp]
\centering
\scriptsize

\setlength{\tabcolsep}{6pt}
\renewcommand{\arraystretch}{1.2}
\caption{Inference Time Comparison of Adaptive and Uniform Protection Methods for Per-Image Execution (ms).}
\label{tab:runtime_comparison}
\begin{threeparttable}
\begin{tabular}{lccccccc}
\toprule
\multirow{2}{*}{\textbf{Dataset@Model}} & 
\multirow{2}{*}{\textbf{Unprotected DNN}} & 
\textbf{Uniform Protection} & 
\multicolumn{2}{c}{\textbf{Adaptive Protection}} & 
\multicolumn{2}{c}{\textbf{Comparison}} \\
\cmidrule(lr){3-3} \cmidrule(lr){4-5} \cmidrule(lr){6-7}
& & \textbf{ABFT Protection} & \textbf{GNN Predictor} & \textbf{ABFT Protection} & \textbf{\makecell{Total Runtime\\Reduction\tnote{1}}} & \textbf{\makecell{Protection Overhead\\Reduction\tnote{2}}} \\
\midrule
MNIST@LeNet & 1.164 & 0.179 & 0.047 & 0.068 & 4.77\% & 35.75\% \\
CIFAR10@AlexNet & 1.424 & 0.214 & 0.057 & 0.101 & 3.42\% & 26.17\% \\
UC Merced@AlexNet & 2.498 & 0.352 & 0.096 & 0.16 & 3.34\% & 27.27\% \\
DAC SDC@YOLO & 11.646 & 1.621 & 0.31 & 0.565 & 5.62\% & 46.02\% \\
ImageNet@ViT-B & 48.009 & 2.735 & 0.431 & 0.74 & 3.08\% & 57.18\% \\
\bottomrule
\end{tabular}

\begin{tablenotes}
\footnotesize

\item[1] $Total\ Runtime\ Reduction = \left ( 1 - \frac{Unprotected\ DNN +  Adaptive\ GNN\ Predictor +  Adaptive\ ABFT\ Protection}{Unprotected\ DNN +  Uniform\ ABFT\ Protection}  \right ) \times 100\%.$

\item[2] $Protection\ Overhead\ Reduction = \left ( 1 -  \frac{Adaptive\ GNN\ Predictor +  Adaptive\ ABFT\ Protection}{Uniform\ ABFT\ Protection} \right )  \times  100\%.$

\end{tablenotes}
\end{threeparttable}
\end{table*}

To demonstrate the time efficiency of the proposed fault tolerance strategy, we compare the runtime of the proposed fault tolerance strategy with the baseline Uniform Protection method during per-image inference across multiple mainstream datasets and CNN models under single fault injection, using a single PH402 SKU 200 GPU card on a server. The results are shown in Table \ref{tab:runtime_comparison}. Experimental results show that the proposed adaptive protection method reduces the total running time by 5.62\% compared to the Uniform protection baseline method.  Regarding the runtime overhead of the protection mechanism itself, the proposed method achieves a significant reduction ranging from 26.17\% to 57.18\%. In terms of runtime composition,  the ABFT protection algorithm introduces additional overhead around 10\% \cite{zhao2020ft, liang2024light, wu2023anatomy}, depending on the target neural network model. Typically, the normalized overhead of the predictors for larger neural network models is much less than that for smaller neural network models. This is as expected because the predictors take network operators as the modeling granularity and the overhead of the predictor increases little with the scale of the target neural network models. Compared to overall neural network execution, the GNN prediction overhead is negligible, especially for larger neural network models.

\subsection{Evaluation of Different Batch Processing Optimizations}

To optimize performance under batch processing, we further propose a vulnerability-aware dynamic batching strategy. We fully leverage the low-overhead GNN predictor to implement a pipelined preprocessing for batch prediction. Specifically, before being fed into the DNN, the system performs vulnerability pre-screening on samples. Based on the prediction results, the scheduler prioritizes aggregating samples labeled as non-vulnerable and vulnerable into distinct batches, such that each batch contains samples of the same vulnerability level to ensure efficient batch processing. Since the preprocessing is usually much faster than the normal neural network processing, it will not affect the performance of the batch processing system when the preprocessing and the batch processing are organized as a unified pipeline.

To evaluate the performance of batch optimization,  we use a YOLO model trained on the DAC SDC dataset as an example. The experiment compares the additional average latency introduced per image when executing fault-tolerant protection  under single-fault injection at different batch sizes, with and without batch protection optimization.  The Without Batch Opt refers to standard batch processing without input reorganization. The With Batch Opt refers to the processing with input reordering and it has inputs split into vulnerable queues and non-vulnerable queues. The experimental results are presented in Table~\ref{tab:batch}. The results demonstrate that the With Batch Opt strategy reduces the average runtime overhead.

\begin{table}[htbp]
\centering
\scriptsize

\setlength{\tabcolsep}{5pt}
\renewcommand{\arraystretch}{1.2}
\caption{Average latency comparison of different batch processing strategies (ms).}
\label{tab:batch}
\begin{tabular}{cccc} 
\toprule
\textbf{Batch Size}  &\textbf{Without Batch Opt.} & \textbf{With Batch Opt.} \\
\midrule
\addlinespace
1 & 0.875 & 0.875 \\ 
64 & 0.074 & 0.06 \\
128 & 0.067 & 0.051 \\
\bottomrule
\end{tabular}
\end{table}

\subsection{Vulnerability Predictor Evaluation}

\subsubsection{Predictor Accuracy} 
\begin{figure}[!t]
\centering

\vspace{-1em}
\includegraphics[width=0.45\textwidth]{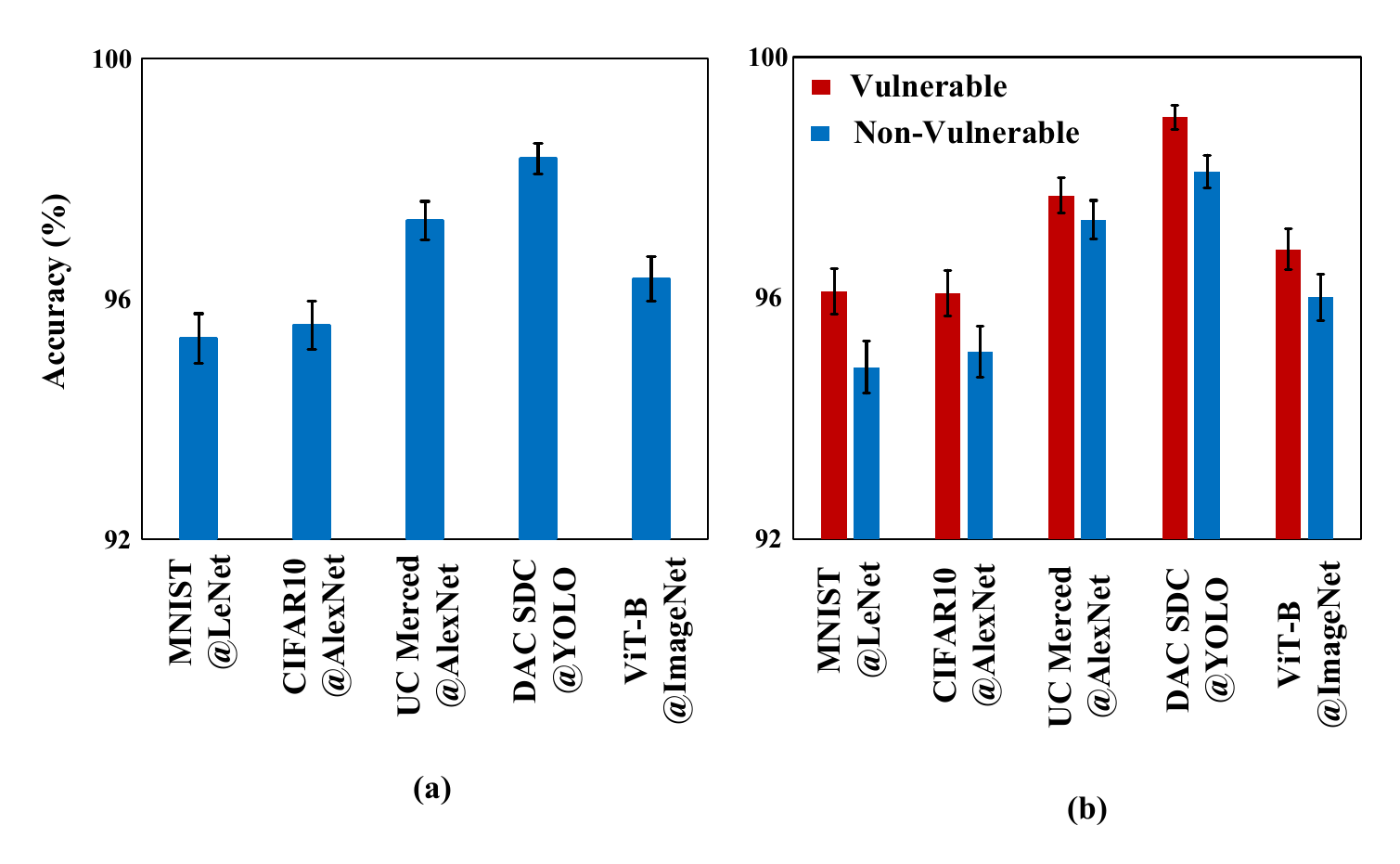}
\vspace{-1em}
\caption{(a) Accuracy of the vulnerability predictor on different datasets and neural network models. (b) Probability of correct classification for different classification labels.}
\label{fig:img4}
\end{figure}

Fig. \ref{fig:img4}(a) compares the accuracy of the GNN vulnerability predictor under five different datasets and neural network models.  The results show that while classification performance varies across models and datasets, the predictor generally achieves impressive results. Notably, it achieves over 95\%  accuracy on multiple tasks, and even exceeds higher standards on the YOLO task of the DAC SDC dataset, demonstrating its robust classification capabilities.  

To gain more insight into the classification performance, Fig. \ref{fig:img4}(b) depicts the correct classification probabilities for vulnerable and non-vulnerable labels. As can be seen from the figure, the classification accuracy for vulnerable node tasks is generally higher than that for non-vulnerable node tasks. This is mainly because vulnerable node tasks have a more significant impact on the fault tolerance performance of the overall deep learning model. Specifically, vulnerable node tasks require stricter fault-tolerant strategies.  If a vulnerable task is misclassified as a non-vulnerable task, it will be subject to a more lenient fault-tolerant strategy, resulting in insufficient protection of the deep learning model and affecting overall performance. Therefore, the optimization goal of the vulnerability predictor is to maximize the classification accuracy for vulnerable tasks and thus enhance the fault-tolerant performance of the entire system.

\subsubsection{Predictor Computational Complexity}   Unlike the complex target neural network model, the vulnerability predictor requires minimal computational resources, as shown in Table \ref{tab:ABFT-def}. The Multiply–Accumulate Operations (MACs) of the target neural network model and the vulnerability predictor model are tested by calling the common PyTorch-OpCounter (thop)\cite{thop}. Provided that the designed vulnerability predictor model remains swift and resource-efficient, any overhead it introduces could be offset by more relaxed fault-tolerant designs for inputs or model components with lower sensitivity. Therefore, it offers both high prediction accuracy and efficient operation even under limited resource conditions.

\begin{table}[ht]\centering
\renewcommand\arraystretch{1.3}
\caption{\# of MACs of target neural network models and the vulnerability predictor.}

\label{tab:ABFT-def}
\scriptsize
\setlength{\tabcolsep}{1mm}{\begin{tabular}{c c c}
\hline
      Dataset@Model        & Target Neural Network & Vulnerability Predictor \\\hline
MNIST@LeNet  & 2.82e5  &    8.85e4                     \\
CIFAR10@AlexNet & 6.27e7   &   1.56e5                      \\
UC Merced@AlexNet &  7.87e8   &  1.56e5                       \\
DAC SDC@YOLO & 1.99e9      &   9.33e5                       \\
ImageNet@ViT-B &  1.69e10     &   1.77e6                       \\\hline

\end{tabular}}
\end{table}

\subsubsection{Analysis of Different Vulnerability Classification Granularities}

To enable more fine-grained selective protection, we set more vulnerability classification categories. As the number of vulnerability classification categories needs to match the corresponding fault-tolerant design approaches and the matching can be complex when more categories are utilized, we only explore two-category and three-category setups in this experiment. Specifically, we utilize the approximate ABFT proposed in \cite{xue2023approxabft} and set different error thresholds to match the two-category and three-category vulnerability setups. Strict ABFT is used for the neural network components classified as the highest vulnerability level and no protection is applied to the neural network components classified as the lowest vulnerability level, which is consistent with the two-category setup. For  neural network components with a moderate vulnerability level, we raise the ABFT error threshold to 10.

\begin{figure}[!t]
\centering
\includegraphics[width=0.48\textwidth]{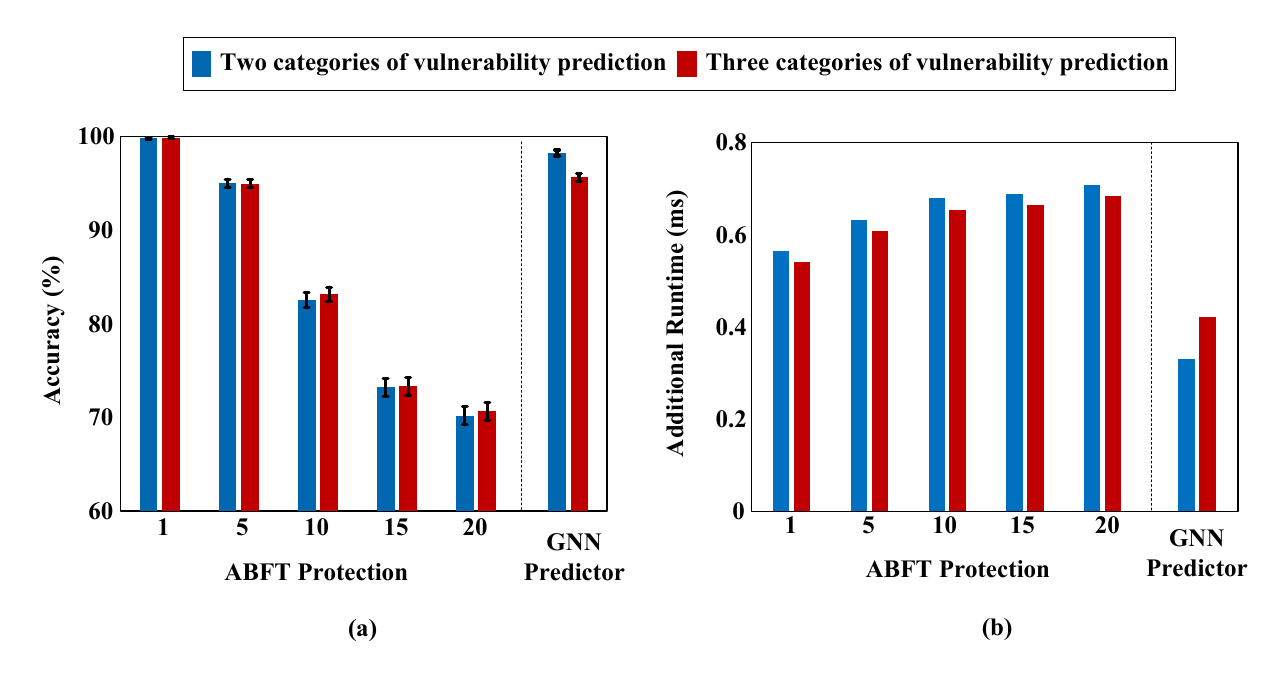}
\vspace{-1em}
\caption{(a) Accuracy of the GNN predictor and the model accuracy under ABFT protection. (b) Additional runtime overhead introduced by the GNN predictor and the ABFT protection algorithm during single-image inference.}
\label{fig:comp_3class}
\end{figure}

Taking the YOLO model on the DAC SDC dataset as an example, Fig. \ref{fig:comp_3class}(a) compares the accuracy of the GNN predictor and the model accuracy with ABFT protection  under two-category and three-category vulnerability classification settings. Fig. \ref{fig:comp_3class}(b) compares the additional runtime overhead introduced by the GNN predictor and ABFT protection algorithm when running single-image inference under two-category and three-category vulnerability classification settings. The results show that two-category classification outperforms the three-category classification for GNN predictor classification accuracy. This is primarily because the finer-grained task increases the complexity and recognition difficulty for the GNN. However, the DNN model accuracy under ABFT protection guided by three-category classification is slightly higher than that of the two-category setting. This indicates that while three-category classification reduces the accuracy of the GNN predictor, it improves the overall accuracy of the DNN model through finer-grained ABFT protection.  For runtime overhead, the ABFT protection algorithm for three-category classification reduces costs by an average of 3.85\% compared to two-category classification. This is mainly because fine-grained classification allows for relaxed ABFT protection on medium-vulnerability components, reducing unnecessary protection in the DNN. However, due to the increased complexity of the GNN classification task, the overhead of the GNN predictor itself increases slightly in the three-class setting.

\subsubsection{Comparison with ML-based vulnerability predictor} 
\begin{figure}[!t]
\centering
\includegraphics[width=0.45\textwidth]{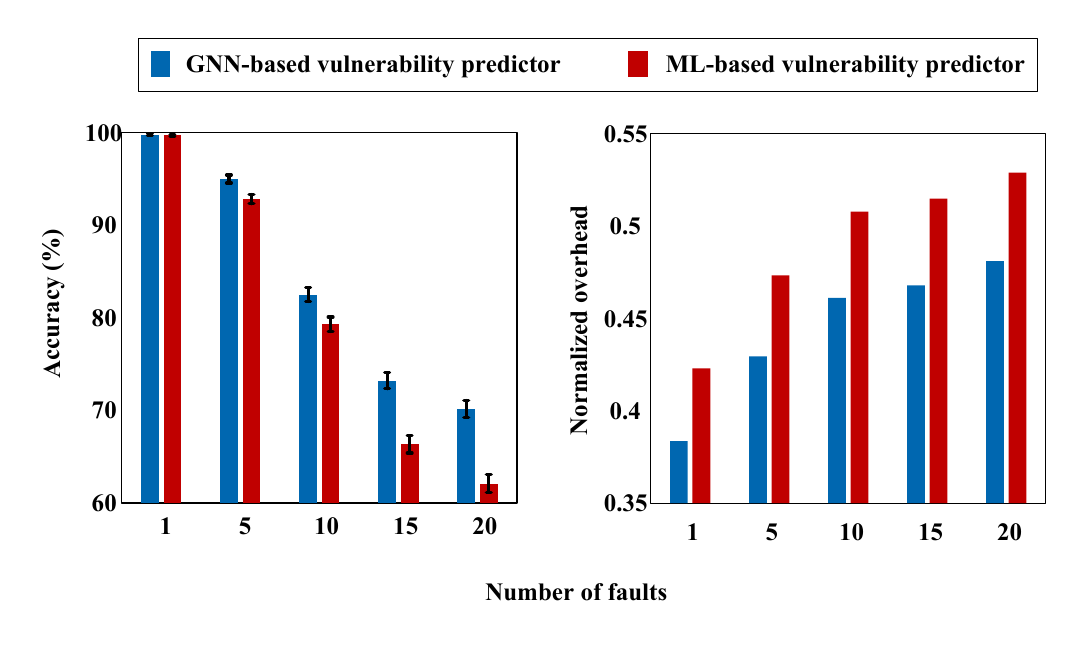}
\vspace{-1em}
\caption{Model accuracy and computational overhead comparison when using different vulnerability predictors on the YOLO model of the DAC SDC dataset. }
\label{fig:comp_ml}
\end{figure}

To demonstrate the advantages of the proposed vulnerability predictor, in Fig. \ref{fig:comp_ml}, we comprehensively compare the proposed GNN vulnerability-aware adaptive fault-tolerant approach with machine learning (ML)-based vulnerability prediction approaches on the YOLO model of the DAC SDC dataset in terms of accuracy and computational overhead. The computational overhead is normalized to that of uniform protection. Specifically, for ML-based vulnerability predictors, we test three classic algorithms: support vector machine (SVM)\cite{platt1998sequential}, random forest (RandomForest)\cite{breiman2001random}, and gradient boosted decision tree (GBDT)\cite{friedman2001greedy}. After comprehensive evaluation, we select the GBDT algorithm with the best performance as the benchmark for the comparative experiment.

The experimental results show that compared with the ML-based vulnerability prediction approach, our proposed GNN-based approach has significant advantages in both accuracy and computational overhead. This advantage is mainly attributed to the robust graph-structured data processing capabilities and global vulnerability perception capabilities of the GNN, which enables it to more effectively capture and utilize the complex fault dependencies and characteristic patterns of target input images and neural network models, thereby achieving more accurate and efficient prediction of vulnerabilities.

\subsubsection{Offline Time Evaluation of  Dataset Preparation and GNN Predictor Training} Table \ref{tab:offline_preparation_time} quantifies the offline time cost of constructing vulnerability datasets and training GNN predictors under different tasks. The results show significant differences in time costs due to varying task complexities. For vulnerability dataset generation, the ViT-B model on the ImageNet dataset takes the longest time, while the AlexNet model on the UC Merced dataset requires only about 1.899 minutes. This is primarily influenced by the original dataset scale and the complexity of the model architecture. For GNN predictor training, the time consumption per epoch exhibits a similar trend. The ViT-B model on the ImageNet dataset requires processing the largest vulnerability graph, thus requiring the longest time for each training epoch. Overall, the time cost of the offline phase mainly depends on the size of the original dataset and the complexity of the model.

\begin{table}[htbp]
  \centering
  \scriptsize
  
  \setlength{\tabcolsep}{5pt}
  \renewcommand{\arraystretch}{1.2}
  \caption{Time Evaluation of Offline Vulnerability Dataset Preparation and GNN Predictor Training.}
  \label{tab:offline_preparation_time}
  \begin{tabular}{lcc}
    \toprule
    \textbf{Dataset@Model} & 
    \makecell{\textbf{Vulnerability Dataset} \\ \textbf{Generation Time (min)}} & 
    \makecell{\textbf{GNN Training} \\ \textbf{ Time Per Epoch (s)}} \\
    \midrule
    MNIST@LeNet         & 49.779  & 1.494  \\
    CIFAR10@AlexNet     & 51.105  & 2.063  \\
    UC Merced@AlexNet   & 1.899   & 0.118  \\
    DAC SDC@YOLO        & 270.249 & 22.08 \\
    ImageNet@ViT-B      & 366.5 & 88.1 \\
    \bottomrule
  \end{tabular}
\end{table}

\section{Conclusion}

In this work, we propose an adaptive fault-tolerant design strategy for neural network processing. It leverages a lightweight GNN model to characterize the soft error vulnerability variation over different inputs and neural network components at runtime. With the predictor, it can adjust the fault-tolerant approaches dynamically to conduct just enough fault protection based on the changing neural network vulnerability factors to ensure cost-effective protection. In addition, it is orthogonal to classical fault-tolerant approaches, and can be integrated with them to suit the various operation environments. Experiments demonstrate that the proposed adaptive soft error protection strategy can significantly optimize classic fault-tolerant approaches. Compared with static selective protection, the computational overhead can be reduced by 42.12\% on average, without affecting the fault-tolerant capability.

\bibliographystyle{ieeetr}
\bibliography{ref}

\begin{IEEEbiography}[{\includegraphics[width=1in,height=1.25in,clip,keepaspectratio]{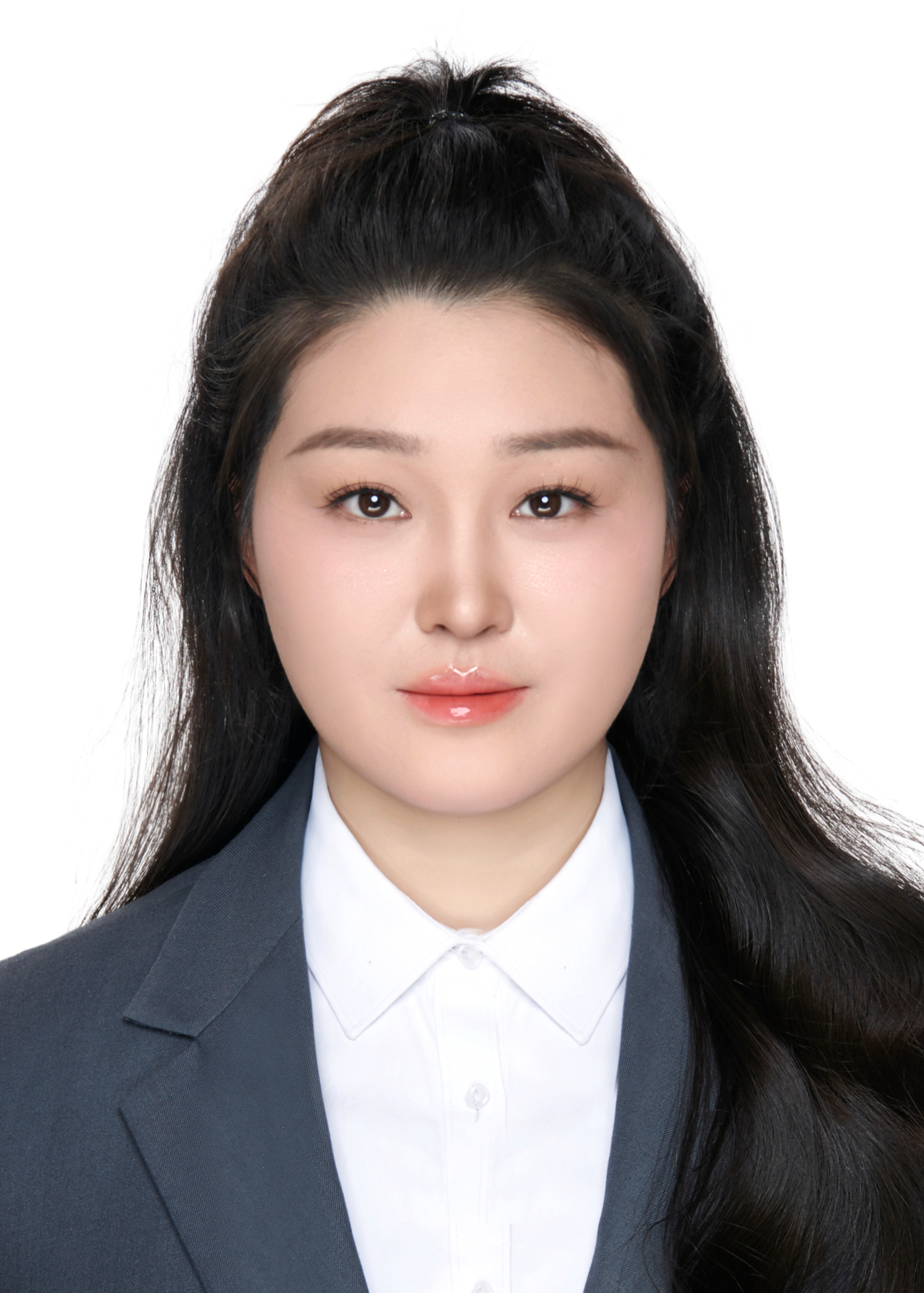}}]{Xinghua Xue} 
received the B.S. degree in electronic and information engineering from Shanxi University, Shanxi, China, in 2017, and M.S. degree in IC engineering from Nankai University, Tianjin, China, and Ph.D degree in Institute of Computing Technology (ICT), Chinese Academy of Sciences(CAS), Beijing, China. Her research interests include fault-tolerant architecture design and robust deep learning.
\end{IEEEbiography}
\vspace{-5 mm} 

\begin{IEEEbiography}[{\includegraphics[width=1in,height=1.25in,clip,keepaspectratio]{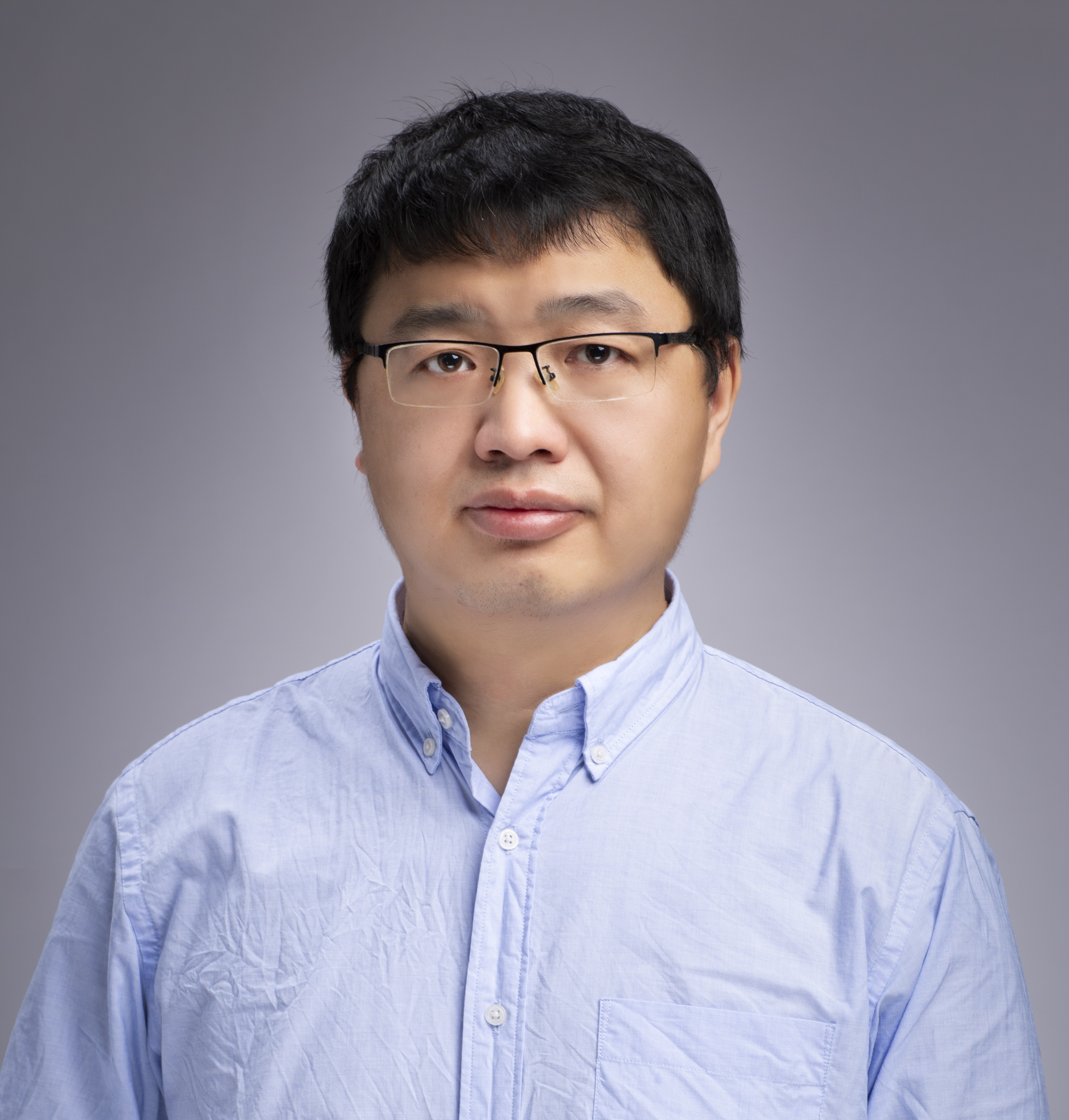}}]{Cheng Liu}
 	received the B.Eng. and M.Eng. degrees from the Harbin Institute of Technology, Harbin, China in 2007 and 2009 respectively, and the Ph.D. degree from the University of Hong Kong in 2016. Currently, he is an associate professor in Institute of Computing Technology (ICT), Chinese Academy of Sciences (CAS). His research interests include FPGA-based reconfigurable computing, fault-tolerant computing, and custom computing.

\end{IEEEbiography}
\vspace{-5 mm} 

\begin{IEEEbiography}[{\includegraphics[width=1in,height=1.25in,clip,keepaspectratio]{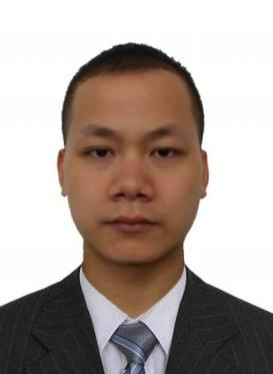}}]{Feng Min}
	received the B.S. and M.S. degrees in Guilin University of electronic technology, Guilin, China, in 2014 and 2017, respectively, and the Ph.D. degree in computer science from the Institute of Computing Technology (ICT), Chinese Academy of Sciences (CAS), Beijing, China, in 2022. He is currently an Assistant Professor with ICT, CAS. His current research interests include computer architecture, domain specific accelerator, VLSI design, and approximate computing.
\end{IEEEbiography}
\vspace{-5 mm} 

\begin{IEEEbiography}[{\includegraphics[width=1in,height=1.25in,clip,keepaspectratio]{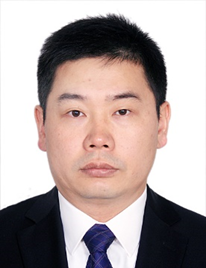}}]{Hui Li}
serves as Director of the National Engineering Research Center for Information Storage and concurrently holds the position of General Manager at Jinan Inspur Data Technology Co., Ltd. He has led two national-level research projects. He was awarded the First Prize for Scientific and Technological Progress in Shandong Province. He specializes in the research and development of storage systems and storage devices.
\end{IEEEbiography}
\vspace{-5 mm} 

\begin{IEEEbiography}[{\includegraphics[width=1in,height=1.25in,clip,keepaspectratio]{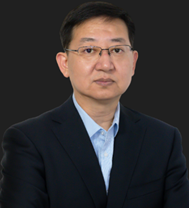}}]{Kai Zhang} 
serves as Director of Cloud Storage Director position at Jinan Inspur Data Technology Co., Ltd. He has been awarded the Provincial First Prize for Scientific and Technological Progress. He holds five authorized national invention patents. His research focuses on the research and development of distributed storage technologies and products.
\end{IEEEbiography}
\vspace{-5 mm} 

\begin{IEEEbiography}[{\includegraphics[width=1in,height=1.25in,clip,keepaspectratio]{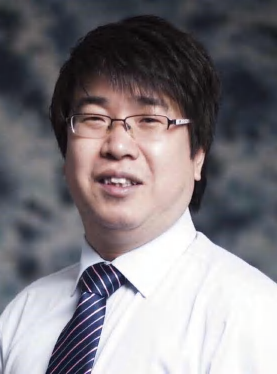}}]{Yinhe Han}
     (Senior Member, IEEE) received the M.S. and Ph.D. degrees in computer science from the Institute of Computing Technology(ICT), Chinese Academy of Sciences(CAS), in2003 and 2006, respectively. He is currently a Professor at ICT, CAS. His main research interests are microprocessor design, integrated circuit design, and computer architecture.

\end{IEEEbiography}
\vspace{-5 mm}

\vfill
\end{document}